\documentclass{article}
\usepackage{graphicx}
\usepackage[main, final]{neurips_2026}

\makeatletter
\AtBeginDocument{%
  \renewcommand{\@noticestring}{}%
}
\makeatother

\usepackage{natbib}
\usepackage{array}
\usepackage{hyperref}
\usepackage{amsmath}
\usepackage{algorithm}
\usepackage{algorithmic}
\usepackage{multirow}
\usepackage{wrapfig}
\usepackage{caption}
\usepackage{enumitem}
\usepackage{caption}
\usepackage{titlesec}
\titlespacing{\section}{0pt}{6pt}{4pt}

\titlespacing{\subsection}{0pt}{5pt}{3pt}

\titlespacing{\subsubsection}{0pt}{4pt}{2pt}
\setlist[itemize]{topsep=2pt, partopsep=0pt, itemsep=1pt, leftmargin=15pt}
\setlist[enumerate]{topsep=2pt, partopsep=0pt, itemsep=1pt, leftmargin=15pt}

\usepackage[utf8]{inputenc} 
\usepackage[T1]{fontenc}    
\usepackage{hyperref}       
\usepackage{url}            
\usepackage{booktabs}       
\usepackage{amsfonts}       
\usepackage{nicefrac}       
\usepackage{microtype}      
\usepackage{xcolor}         
\title{SymbolLKG: Towards Verifiable Logical Reasoning via Logical Knowledge Graph and Symbolic Solvers}

\author{
Haizhao Fan\\ Shanghai JiaoTong University\\
\And
Yuchi Xiong\\ Shanghai JiaoTong University\\
\And
Jize Wang \\ Shanghai JiaoTong University\\
\AND
Xinping Guan\\ Shanghai JiaoTong University\\
\And
Xinyi Le\\ Shanghai JiaoTong University\\
}
\begin{document}
\setlength{\intextsep}{4pt}
\maketitle

\begin{abstract}
Large Language Models (LLMs) have demonstrated remarkable proficiency in natural language understanding, yet they struggle with strict multi-step reasoning, frequently suffering from hallucinations and inconsistency. Existing solutions like Chain-of-Thought (CoT) lack rigorous verification mechanisms, while standard Retrieval-Augmented Generation (RAG) often misses the complex, structural dependencies inherent in logical tasks. To bridge this gap, we propose a Neuro-Symbolic architecture that integrates a Logical Knowledge Graph (LKG) with dynamic solver routing. Specifically, we introduce an ontology-based LKG that treats logical rules and constraints as first-class topological nodes, enabling explicit modeling of dependencies extracted from text. We further design a Logic Router to dynamically dispatch tasks to the optimal symbolic engine, which is supported by a topology-aware hybrid retrieval mechanism. Experimental results on logical reasoning benchmarks demonstrate that our framework significantly outperforms state-of-the-art prompting and RAG baselines, delivering higher accuracy and verifiable reasoning paths.
\end{abstract}

\section{Introduction}

Logical reasoning is the cornerstone of reliable artificial intelligence, particularly in high-stakes domains where precision is mandatory—healthcare \cite{thirunavukarasu2023large}, law \cite{katz2024gpt}, finance \cite{wu2023bloomberggpt}, and scientific discovery \cite{wang2023scientific}. Driven by their natural language understanding capabilities, Large Language Models (LLMs) are increasingly deployed for complex tasks in these fields. However, the probabilistic nature of their ``next-token prediction'' architecture fundamentally limits the rigorous, multi-step logic required for strict reasoning scenarios.

Reasoning encompasses deductive, inductive, and abductive forms \cite{patil2025advancing}; yet achieving trustworthy logical reasoning, characterized by reliability, interpretability, and adherence to formal rules, remains a fundamental challenge \cite{patil2025advancing}. LLMs tend to generate plausible but incorrect steps and fail to maintain consistency across long contexts \cite{valmeekam2023planning}.

To address these issues, prompting methods such as Chain-of-Thought (CoT) \cite{wei2022chain} and Tree-of-Thoughts (ToT) \cite{Yao2023tree} decompose complex problems, and Retrieval-Augmented Generation (RAG) \cite{Lewis2020retrieval} grounds responses in external data. However, these methods lack rigorous verification: CoT can propagate errors without symbolic grounding \cite{Gao2023pal}, and standard RAG misses the directional dependencies inherent in logical rules.

In contrast, Knowledge Graphs (KGs) provide structured models of entities and relationships \cite{hogan2021knowledge,ji2021survey}, supporting rigorous formal reasoning and interpretability \cite{patil2025advancing}; but symbolic systems traditionally struggle with scalability on open-ended unstructured data. Integrating KGs with LLMs has thus emerged as a promising direction \cite{pan2024unifying}: Think-on-Graph \cite{sun2023think} and RoG \cite{luo2023reasoning} guide LLMs along graph paths; GraphRAG \cite{edge2024local} structures retrieval from local contexts to global summarization; LightRAG \cite{guo2024lightrag} introduces a dual-level retrieval mechanism capturing both specific entities and conceptual themes; and KAG \cite{liang2025kag} pushes neuro-symbolic alignment in professional domains by coupling LLMs with a logical-form-guided engine. However, these approaches typically treat logical rules as unstructured text or implicit edge attributes rather than explicit, computable topological nodes. This representational gap limits the deterministic mapping required for rigorous symbolic verification, often forcing reliance on probabilistic text-to-code generation. Without distinct constraint nodes, analyzing the logical topology to dynamically route problems to the appropriate solver (e.g., Z3 vs. Prover9) also becomes significantly more challenging, restricting adaptability across diverse reasoning tasks.

\begin{wrapfigure}{r}{0.4\textwidth}
  \centering
  \begin{minipage}{\linewidth}
    \includegraphics[width=\linewidth]{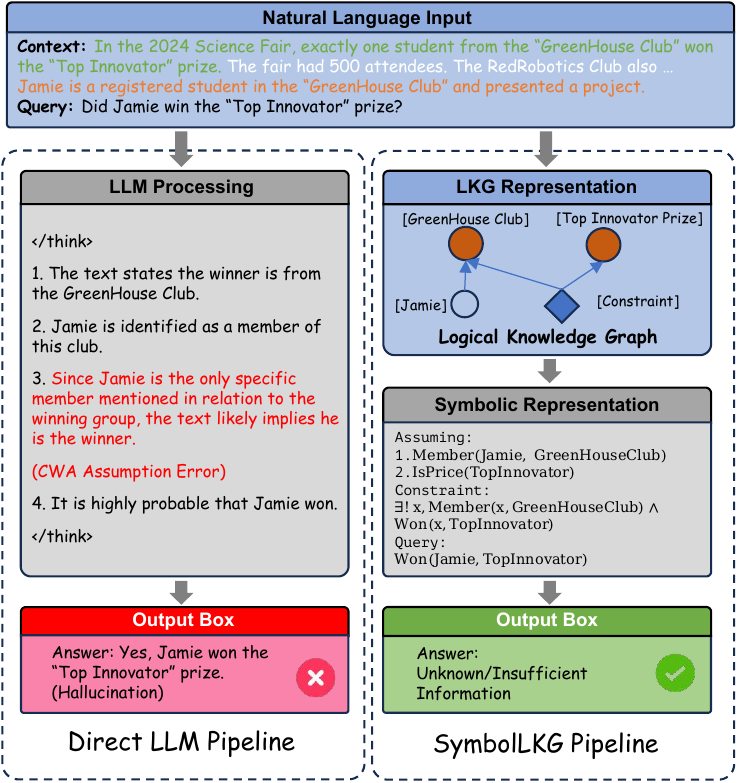} 
    \caption{\textbf{Comparison of reasoning paradigms.} The input highlights constraints in green, noise in grey, and premises in orange. The Direct LLM Pipeline (left) ignores this structure. It makes a ``Closed World Assumption'' (CWA) error and incorrectly guesses Jamie won. The SymbolLKG Pipeline (right) converts text into a Logical Knowledge Graph. It uses symbolic verification to strictly check the premises. The system correctly concludes the answer is ``Unknown''.} 
  \end{minipage}
\end{wrapfigure}

Driven by these insights, we propose a neuro-symbolic framework that integrates Logical Knowledge Graphs (LKG) with external symbolic solvers, treating logical rules and constraints as first-class nodes. This combines the natural language understanding of LLMs for parsing problems into a structured graph with the deterministic power of external solvers for the actual reasoning.

Our main contributions are summarized as follows:
\begin{itemize}

\item \textbf{SymbolLKG Framework:} 
We propose SymbolLKG, a novel neuro-symbolic framework that grounds LLM reasoning via Logical Knowledge Graphs (LKG) and router-guided solvers. The pipeline parses natural language into an LKG, retrieves a topologically connected subgraph, and dynamically routes the problem to the optimal symbolic solver, bridging neural semantic flexibility and symbolic deterministic precision.

\item \textbf{Topology-Aware Logical Knowledge Graph:} 
We introduce a dynamic LKG architecture that treats Rules and Constraints as independent nodes, using a ``Schema-on-Read'' strategy to build Concept-Entity hierarchies from context. This topology enables hybrid retrieval combining vector search with graph traversal, refined via LLM pruning into a \textit{Logical Hull}—a complete yet minimal subgraph of essential premises.

\item \textbf{Adaptive Logic Routing and Self-Refining Translation:} 
We implement a Logic Router that analyzes problem structure to select the optimal symbolic tool (e.g., Z3 for math, Prover9 for logic), with a feedback loop where the LLM uses solver error traces to refine its generated code, improving reliability on complex logical tasks.
\end{itemize}

We evaluate our framework on standard logical reasoning and multi-hop QA datasets. Experimental results demonstrate that our approach significantly outperforms standard LLM prompting and baseline RAG methods, validating the efficacy of combining structured knowledge graphs with deterministic solvers for trustworthy reasoning.

\section{Related Works}
\textbf{Large Language Models for Logical Reasoning.} 
Large Language Models show strong performance in reasoning tasks. Techniques like CoT \cite{wei2022chain} and Zero-Shot CoT \cite{kojima2022large} are now standard for generating multi-step reasoning. To improve reliability, methods such as Self-Consistency \cite{wang2022self} and ToT \cite{Yao2023tree} explore multiple reasoning paths. Despite these advances, pure neural models still suffer from hallucinations and ``unfaithful'' explanations, meaning their generated reasons often do not match their actual decision process \cite{Turpin2023language}. Furthermore, Transformers struggle with compositional generalization, failing on multi-hop deduction problems outside their training data \cite{Dziri2023faith,nye2021show}. Self-Correction is also difficult because models rarely catch their own logical errors without external feedback \cite{Huang2023large}. This highlights the need for robust external verification mechanisms.

\textbf{Neuro-Symbolic AI and Auto-formalization.}
Neuro-symbolic AI combines neural flexibility with symbolic rigor. A key approach is auto-formalization, where LLMs translate text into formal code \cite{Chen2022program,Gao2023pal} or logic rules \cite{Lyu2023faithful}. Systems like Logic-LM \cite{Pan2023logic} and LINC \cite{Olausson2023linc} use external solvers to execute these rules. However, most existing systems use a static strategy, assigning one solver (e.g., Z3 or Prover9) to an entire dataset. This rigid method fails on complex queries mixing arithmetic and relational logic. As shown by \cite{Lam2024closer}, tool effectiveness depends heavily on the specific structure of each problem. Our SymbolLKG solves this with Dynamic Solver Routing. Instead of using dataset-level rules, our system acts as a controller, analyzing each query at runtime to select the best solver.

\textbf{Retrieval-Augmented Generation for Structured Knowledge.} Retrieval-augmented generation (RAG) reduces hallucinations by using external documents \cite{Lewis2020retrieval,Guu2020retrieval}. Standard RAG relies on dense vector similarity to fetch relevant context. However, for logical reasoning tasks, this often leads to ``semantic drift''—where the system retrieves semantically similar but logically irrelevant documents. While good for simple facts, standard retrieval fails at multi-hop reasoning, which requires following logical chains rather than just matching similar words. Recent works use Knowledge Graphs to solve this. Methods like Think-on-Graph \cite{sun2023think}, RoG \cite{luo2023reasoning}, StructGPT \cite{jiang2023structgpt} and GraphRAG \cite{edge2024local}  guide LLMs through graph structures. Others, like Selection-Inference \cite{creswell2022selection}, break reasoning into smaller steps. However, these methods usually treat KGs as static lists of entities and ignore complex logical rules. We propose a Logical Knowledge Graph (LKG) that treats rules and constraints as primary nodes. By combining vector search with Logical Hull Expansion, we ensure implicit axioms are retrieved for the symbolic solver, bridging the gap between semantic search and logical deduction.

\section{Methodology}

\begin{figure}[htbp] 
    \centering
    \includegraphics[width=\textwidth]{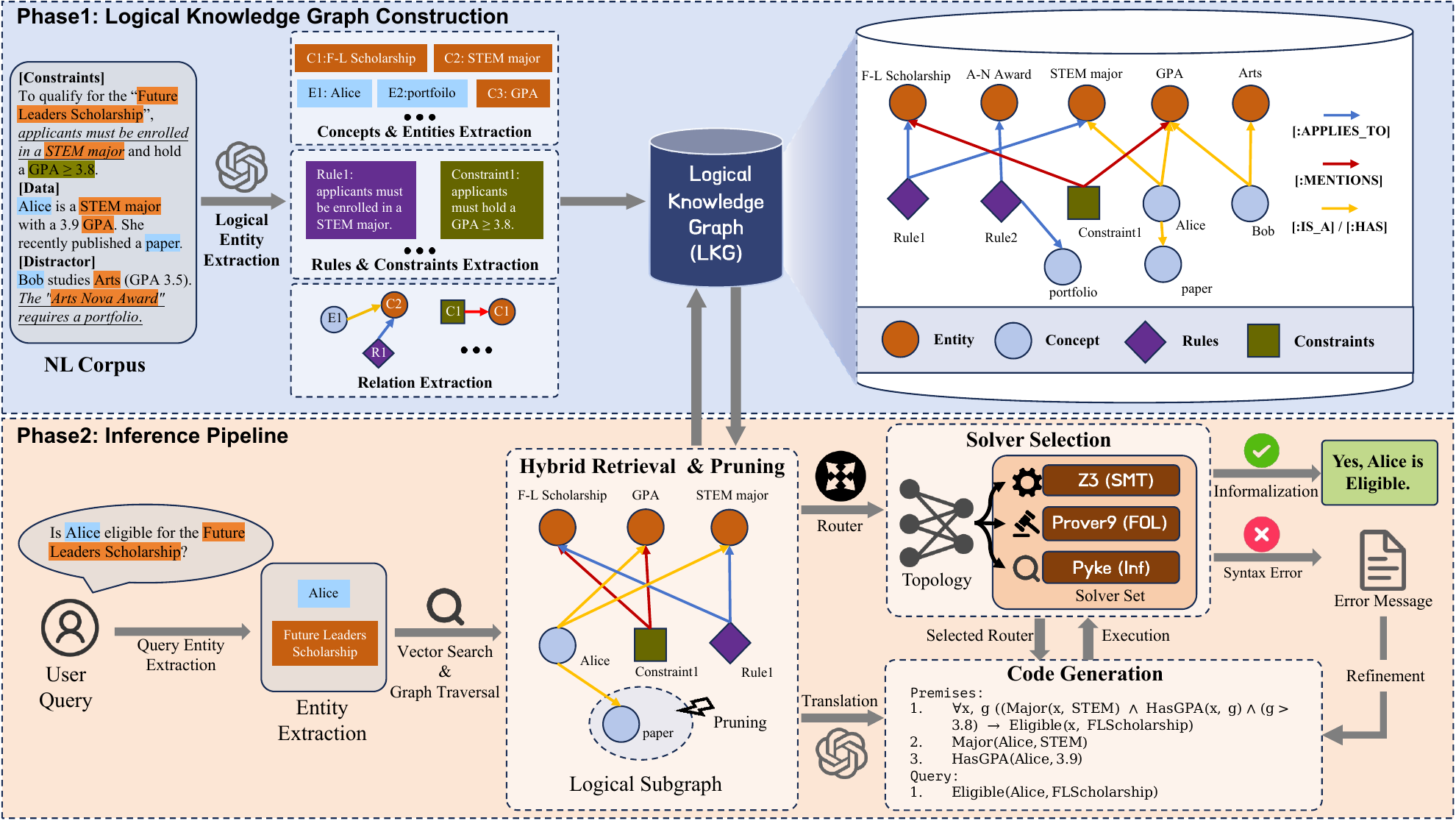}

    \caption{\textbf{The architecture of the SymbolLKG framework.} The pipeline consists of two phases: \textbf{Phase 1 (LKG Construction)} transforms text into an LKG, establishing a dynamic Concept-Entity hierarchy alongside reified Rule and Constraint nodes. 
    \textbf{Phase 2 (Inference)} employs a hybrid retrieval strategy (combining vector search with graph traversal) to extract a context-aware subgraph. Then a topology-aware router dispatch it to the optimal symbolic solver for verifiable execution.}
    
    \label{fig:pipeline}
\end{figure}

We propose the \textbf{Symbol}ic-Cognitive \textbf{L}ogical \textbf{K}nowledge \textbf{G}raph (\textbf{SymbolLKG}) framework, a neuro-symbolic architecture (\hyperref[fig:pipeline]{Figure 2}) that maps unstructured text into verifiable reasoning paths. The workflow begins at the top with the Logical Knowledge Graph Construction phase, where an LLM parses the natural language corpus to extract entities, concepts, rules, and constraints into a structured LKG. The process flows into the Inference Pipeline, which first employs Hybrid Retrieval and Pruning to isolate a context-aware subgraph containing only relevant premises. Next, an Adaptive Logic Router analyzes the subgraph's topology to pick the best solver, passing the problem to the Code Generation module for symbolic execution and self-refining verification.

\subsection{Logical Knowledge Graph Construction}

Phase 1 begins by processing the natural language corpus through an LLM-based extraction pipeline. As illustrated in the ``LKG Construction'' panel of \hyperref[fig:pipeline]{Figure 2}, we adopt an Open Information Extraction (OpenIE) paradigm to handle the diversity of natural language. We formally define the LKG as a directed, heterogeneous multigraph $G = (V, E, \mathcal{A}, \mathcal{T})$, where $V$ is the set of nodes, $E$ is the set of edges, $\mathcal{A}$ is a set of attributes, and $\mathcal{T}$ represents the node types. The vertex set $V$ is partitioned into four disjoint subsets, as shown in \hyperref[tab:lkg_nodes]{Table1}.

First, we extract Concepts and Entities using a dynamic ``Schema-on-Read'' strategy. Unlike traditional graphs with rigid categories, our system generates Concept nodes (e.g., ``Student'') and Entity nodes (e.g., ``Alice'') based on their specific roles in the text. This approach ensures that the graph captures the precise domain of discourse needed for logical reasoning.

Next, we identify logical dependencies and instantiate them as Rule and Constraint nodes. We treat logic as ``first-class citizens'' rather than simple text attributes. For example, a restriction like $\text{GPA} > 3.8$ is extracted as a specific Constraint node, while an implication like ``If A then B'' becomes a Rule node. These nodes are then connected to their respective entities via relation edges.

Finally, these components are assembled into the Logical Knowledge Graph (LKG). To ensure consistency, we assign a unique ID to each node using a hash function. This step merges variations like ``The blue book'' and ``Blue Book'' into a single node, preventing duplicates.

This design offers three key advantages. First, it makes logic semantically addressable. Rules get their own vector embeddings $\mathbf{h}_{v} \in \mathbb{R}^d$, so the system can find ``seating restrictions'' directly via semantic search. Second, it separates the T-Box (logical rules) from the A-Box (assertions) \cite{baader2003description}. This keeps the reasoning structure stable even when specific entity data changes. Third, the Concept nodes in our framework are extracted directly from the corpus based on their logical role in the specific problem context, rather than mapping to a universal ontology. They define the \textbf{Domains of Discourse} for logical quantifiers (e.g., $\forall x \in \text{Students}$), enabling the direct translation of natural language into verifiable First-Order Logic (FOL). This ``Schema-on-Read'' approach ensures that symbolic solvers operate on precise, problem-relevant subsets rather than generic classes.

\subsection{Hybrid Retrieval and Pruning}

After the construction of the LKG, the inference pipeline begins by setting a precise entry point for the user query. As shown in the ``Hybrid Retrieval'' module (\hyperref[fig:pipeline]{Figure 2}), we first perform Query Entity Extraction to ground the search. The system parses the natural language query to find specific target entities (e.g., ``Alice'' or ``Future Leaders Scholarship'') before any retrieval occurs.

These extracted entities, combined with the original query, drive the identification of Anchor Nodes. We employ a dense embedding model (BGE-M3 \cite{chen2024bge}) to encode the user query and calculate the cosine similarity between query embedding $\textbf{v}_q$ and all node embeddings. Simultaneously, we perform exact matching against the names of the entities extracted in the previous step. This dual approach selects a set of anchors $S_{\text{anchor}}$ defined as:
\begin{equation}
    \begin{split}
         S_{\text{anchor}} &= \{ v \in V \mid \cos(\mathbf{v}_q, \mathbf{h}_v) > \tau_{sim} \} \\
&\cup \{ v \in V_E \mid \text{exactmatch}(v.name, q) \}
    \end{split}
\end{equation}
This strategy captures both the explicit targets mentioned by the user and the abstract logical concepts relevant to the context.

However, semantic search alone is not enough for logic. A rule constraining an entity often lacks shared keywords with that entity. To fix this, we define the \textbf{Logical Hull} of the anchor set. We expand the context by traversing the graph's [:MENTIONS], [:IS\_A] and [:APPLIES\_TO] edges. For each anchor $v \in S_{\text{anchor}}$, we include its k-hop (determined by datasets) neighbors from the rule set $V_R$ or constraint set $V_S$.
\begin{equation}
\begin{split}
    S_{\text{tmp}} &= \{ u \in V_R \cup V_S \mid \exists v \in S_{\text{anchor}}, \text{dist}(u, v) \le k \} \\
    S_{\text{hull}} &= S_{\text{anchor}} \cup S_{\text{tmp}}
\end{split}
\end{equation}
This step ensures Logical Completeness. If an entity is found, its related rules are automatically added. E.g., finding ``Alice'' also fetches the ``AllDifferent'' constraint for her group.

Finally, the expanded set $S_{\text{hull}}$ can be large. To fit the solver's context window, we use an LLM-based pruning module $\Psi$. This module checks if each node is needed to answer $q$ and removes ``distractor'' constraints. The result is $G_{\text{final}} = \Psi(S_{\text{hull}}, q)$, a compact, mathematically bounded context for reasoning.

\begin{table}[t]
\centering
\label{tab:lkg_nodes}
\resizebox{\textwidth}{!}{%
\begin{tabular}{l c >{\raggedright\arraybackslash}p{6.0cm} >{\raggedright\arraybackslash}p{4.5cm} >{\raggedright\arraybackslash}p{5.0cm}}
\toprule
\multicolumn{1}{c}{\textbf{Node Type}} & 
\multicolumn{1}{c}{\textbf{Symbol}} & 
\multicolumn{1}{c}{\textbf{Definition \& Categories (Codebase)}} & 
\multicolumn{1}{c}{\textbf{Key Attributes (Class Fields)}} & 
\multicolumn{1}{c}{\textbf{Role in Reasoning}} \\
\midrule

\textbf{Entity} & $V_E$ & 
Concrete instances implemented as \texttt{Entity} class. \newline
IDs are generated via SHA256 hash of lemmatized name and type. & 
\texttt{name}, \texttt{entity\_type}, \newline \texttt{id} (Canonical Hash), \newline \texttt{properties} & 
Anchors for retrieval. Hash-based IDs ensure uniqueness across the graph. \\
\addlinespace[0.8em]

\textbf{Concept} & $V_C$ & 
Abstract categories implemented as \texttt{Concept} class. & 
\texttt{name}, \texttt{labels} & 
Simple taxonomy nodes supporting type grouping. \\
\addlinespace[0.8em]

\textbf{Rule} & $V_R$ & 
Logical implications implemented as \texttt{Rule} class. & 
\texttt{expression}, \texttt{description}, \newline \texttt{embedding} & 
Stores logic formulas; \texttt{description} allows vector-based retrieval of rules. \\
\addlinespace[0.8em]

\textbf{Constraint} & $V_S$ & 
Restrictions on states, implemented as 4 subclasses: \newline 
1. \textbf{Arithmetic} (Math ops) \newline
2. \textbf{AllDifferent} (Combinatorial) \newline
3. \textbf{Ordering} (Spatial/Temporal relations like 
\textit{left\_of}, \textit{newer\_than}) \newline
4. \textbf{Generic} (Constraints outside the above-listed categories) & 
\texttt{raw\_expression}, \texttt{constraint\_type}, 
\newline
\textit{Specific:} \texttt{relation}, \texttt{operation}, \texttt{value} & 
Defines solution boundaries. \texttt{Ordering} handles both topological and chronological constraints via relation types. \\
\bottomrule
\end{tabular}%
}
\vspace{5pt}
\caption{Node Taxonomy in the Symbolic-Cognitive Logical Knowledge Graph (SymbolLKG). The schema distinguishes between static entities, hierarchical concepts, deductive rules, and arithmetic constraints to facilitate dynamic solver routing.}
\end{table}

\subsection{Solver Selection via Logic Router}

Given the pruned subgraph $G_{\text{final}}$, the framework next determines the best strategy to solve the logical problem. As depicted in the ``Solver Selection'' block of \hyperlink{fig:pipeline}{Figure 2}, we introduce an Adaptive Logic Router to analyze the topology of the subgraph.

Real-world logical problems vary widely, so using a single solver is often inefficient. Research has shown that while Z3 \cite{de2008z3} excels at arithmetic and combinatorial tasks, it is often less intuitive for pure first-order logic compared to Prover9 \cite{Lam2024closer}. To optimize performance, we implement an Adaptive Logic Router. This meta-cognitive module $f_{\text{route}}: (G_{\text{final}}, q) \rightarrow \Omega$ analyzes the subgraph's structure to select the best solver backend.

\paragraph{SMT Solver (Z3) Pathway:} Used when $G_{\text{final}}$ has many $V_S$ nodes (Constraints), especially for arithmetic or global uniqueness (e.g., CSPs). We leverage Z3's DPLL(T) algorithm to efficiently prune large search spaces.

\paragraph{Automated Theorem Prover (Prover9) Pathway:} Selected for strict FOL tasks or proof generation. This runs when the subgraph is mostly $V_R$ nodes (Rules) with logical implications but no numbers.

\paragraph{Pyke Pathway:} Used for direct relational queries (e.g., ``Who is X's father?''). Here, lightweight engines like Pyke are faster and more efficient than heavy solvers.

This adaptive approach ensures robustness. By categorizing the problem, we stop the LLM from trying unreliable arithmetic through token prediction—a known Transformer weakness. Instead, we force the system to delegate computation to symbolic engines designed for mathematical correctness.

\subsection{Code Generation}

Once the optimal solver is determined, the final phase translates the logical subgraph into executable symbolic code. The LLM acts as a compiler, mapping the nodes in $G_{\text{final}}$ directly to solver-specific syntax. For instance, in the Z3 pathway, entities $e \in V_E$ become variables ($x = \text{Int}(x)$); concept nodes $V_C$ set domains ($\text{solver}.\text{add}(0 \le x \le 100)$); and constraint nodes $v_c \in V_S$ become assertions ($\text{solver}.\text{add}(\text{Distinct}(x, y, z))$).

To ensure reliability, we use a self-refining execution loop. As shown in the bottom-right of \hyperlink{fig:pipeline}{Figure 2}, the system captures any error messages during execution. It then feeds this feedback back to the model to refine the code. This cycle repeats until success or timeout.

This proposed methodology effectively transforms the stochastic nature of probabilistic text generation into the precision of deterministic computation. Consequently, the resulting output constitutes a rigorous formal proof or a logically valid assignment, delivering a level of interpretability and systemic trust that remains fundamentally unattainable through standard neural Chain-of-Thought prompting alone.

\section{Experiments}

\subsection{Experimental Settings}

\paragraph{Datasets} To comprehensively evaluate the SymbolLKG framework, we utilize eight diverse benchmarks that span the spectrum from formal symbolic logic to open-domain multi-hop reasoning. We categorize them into two groups based on the evaluation focus:

\paragraph{Logical Reasoning Benchmarks} These datasets validate translation precision and deductive validity. \textbf{FOLIO} \cite{han2024folio} provides premises with FOL ground truths and complex logical structures. \textbf{AR-LSAT} \cite{zhong2022analytical} focuses on analytical reasoning and constraint satisfaction from law exams. \textbf{ProofWriter} \cite{tafjord2021proofwriter} assesses rule-based reasoning over multi-hop chains and proof generation. \textbf{LogicalDeduction} \cite{srivastava2023beyond} uses constraint satisfaction problems for abstract reasoning, while \textbf{ProntoQA} \cite{saparov2022language} evaluates transitive reasoning via synthetic multi-hop deductive chains.

\paragraph{Multi-hop Retrieval \& QA Benchmarks} These benchmarks assess the robustness of LKG Construction and Hybrid Retrieval. \textbf{2WikiMultiHopQA} \cite{ho2020constructing} requires information aggregation across multiple articles with reasoning chains up to 4 hops. \textbf{HotpotQA} \cite{yang2018hotpotqa} tests noise filtration and bridge-entity identification by requiring the system to find supporting facts among irrelevant distractors. \textbf{Musique} \cite{trivedi2022musique} challenges models with highly compositional questions designed to systematically minimize disconnected reasoning shortcuts.

\paragraph{Implementation Details} Our SymbolLKG framework utilizes Llama-3.3-70B-Instruct as the backbone LLM for the extraction and translation modules due to its favorable balance between economic efficiency and inference speed. To ensure a fair comparison, we consistently applied Llama-3.3-70B-Instruct as the backbone model for our primary baseline methods, including IRCoT + HippoRAG. To minimize generation randomness and ensure reproducibility, we set the temperature to 0 for all LLM inference calls. For the symbolic solvers, we employ a hybrid strategy: Z3 is used for constraint satisfaction and arithmetic tasks, Prover9 handles strict first-order logic theorem proving, and Pyke is utilized as a lightweight inference engine for direct relational queries. The hybrid retrieval combines dense embeddings (using BGE-M3 \cite{chen2024bge}) with our topology-aware graph traversal.

\subsection{Logical Reasoning Performance}

\paragraph{Baselines} We compare our framework against two representative settings: \textbf{Standard LLM with CoT:} Uses LLM with Chain-of-Thought prompting to establish a strong proprietary model baseline. \textbf{Logic-LM:} A neuro-symbolic framework that integrates LLMs with symbolic solvers, serving as a direct competitor in the symbolic reasoning domain.
\begin{table}[htbp]
    \centering
    \small\scshape
    \begin{tabular}{l|ccc}
    \toprule
    \textbf{Dataset} & \textbf{CoT} & \textbf{Logic-LM} & \textbf{Ours} \\
    \midrule
    FOLIO             & 70.58 & \textbf{78.92} & 71.39 \\
    AR-LSAT           & 35.06 & 43.04 & \textbf{57.85} \\
    Logical Deduction & 75.25 & 87.63 & \textbf{90.81} \\
    ProntoQA          & 98.79 & 83.20 & \textbf{100.00} \\
    ProofWriter       & 68.11 & \textbf{79.66} & 73.60 \\
    \midrule
    Avg.              & 69.56 & 74.49 & \textbf{78.73}\\
    \bottomrule
    \end{tabular}
    \vspace{5pt}
    \caption{Logical Reasoning Accuracy (\%) across five benchmarks. Methods compared include Standard CoT prompting and the neuro-symbolic baseline Logic-LM.}
    \label{tab:reasoning_results}
\end{table}

\paragraph{Results} \hyperref[tab:reasoning_results]{Table 2} summarizes the accuracy across five benchmarks. SymbolLKG achieves an average of $78.73\%$, outperforming Logic-LM ($74.49\%$) and Standard CoT ($69.56\%$). Two design choices drive its strongest cells: the typed constraint subclasses (Ordering / AllDifferent / Arithmetic) feed a structural signal to the Logic Router, deterministically selecting Z3 on AR-LSAT and yielding $57.85\%$; and the three-attempt self-refining loop, which echoes solver compiler errors back to the LLM, completes ProntoQA's transitive Horn chains at $100.00\%$. SymbolLKG lags Logic-LM on FOLIO and ProofWriter because Logic-LM relies on dataset-specific
prompts, whereas our generalist ontology often demotes free-form quantified premises to Generic rules, weakening the router signal and limiting the recovery scope of compiler-trace feedback.

\subsection{Multi-hop Retrieval Performance}
 
We evaluate the retrieval effectiveness of our system on two challenging multi-hop datasets: 2WikiMultiHopQA, HotpotQA and Musique. This experiment tests the system's ability to identify and retrieve disjointed pieces of evidence required for multi-step reasoning.

\paragraph{Baselines} We categorize our baselines into Single-step and Multi-step retrieval approaches to provide a comprehensive comparison: \textbf{Single-step Retrieval:} We include sparse retrieval (BM25 \cite{robertson1994some}), dense retrieval (Contriever \cite{izacard2021unsupervised}, GTR \cite{ni2022large}, NativeRAG \cite{Lewis2020retrieval}), and advanced structure-aware methods (RAPTOR \cite{sarthi2024raptor}, Proposition \cite{chen2024dense}, HippoRAG \cite{jimenez2024hipporag}). \textbf{Multi-step Retrieval:} We evaluate IRCoT \cite{trivedi2023interleaving} (Interleaving Retrieval with Chain-of-Thought) combined with four different retrievers: BM25, Contriever, NativeRAG, and HippoRAG. This setup allows us to compare our topology-aware retrieval against iterative feedback loops powered by different backbone retrievers.

\paragraph{Metrics} We report \textbf{Recall@k}, defined as the proportion of ground-truth supporting documents found within the top-$k$ retrieved candidates. Given the ground-truth supporting fact set $G_q$ and the top-$k$ retrieved set $R_{q,k}$ for a query $q$, the recall is calculated as:
\begin{equation}
\text{Recall@}k = \frac{1}{|Q|} \sum_{q \in Q} \frac{|R_{q,k} \cap G_q|}{|G_q|}
\end{equation}
We use Recall@2 and Recall@5, which measure the percentage of queries where the top-2 and top-5 retrieved documents contain the complete set of ground-truth supporting facts.

\begin{table*}[htbp]
\label{tab:retrieval_results}
\vskip 0.15in
\begin{center}
\begin{small}
\begin{sc}
\begin{tabular}{ll|cc|cc}
\toprule
\multirow{2}{*}{\textbf{Type}} & \multirow{2}{*}{\textbf{Method}} & \multicolumn{2}{c|}{\textbf{2WikiMultiHopQA}} & \multicolumn{2}{c}{\textbf{HotpotQA}} \\
 & & Recall@2 & Recall@5 & Recall@2 & Recall@5 \\
\midrule
\multirow{7}{*}{Single-step} 
 & BM25              & 51.8 & 61.9 & 55.4 & 72.2 \\
 & Contriever        & 46.6 & 57.5 & 57.2 & 75.5 \\
 & GTR               & 60.2 & 67.9 & 59.4 & 73.3 \\
 & RAPTOR            & 46.3 & 53.8 & 58.1 & 71.2 \\
 & Proposition       & 56.4 & 63.1 & 58.7 & 71.1 \\
 & NativeRAG         & 59.2 & 68.2 & 64.7 & 79.3 \\
 & HippoRAG          & 70.7 & \underline{89.1} & 60.5 & 77.7 \\
\midrule
\multirow{4}{*}{Multi-step} 
 & IRCoT + BM25      & 55.4 & 66.8 & 62.1 & 75.3 \\
 & IRCoT + Contriever& 51.6 & 63.8 & 65.9 & 81.6 \\
 & IRCoT + NativeRAG & 64.1 & 74.4 & \underline{67.9} & 82.0 \\
 & IRCoT + HippoRAG  & \underline{75.3} & \textbf{93.4} & 67.0 & \underline{83.0} \\
\midrule
\textbf{Ours} 
 & \textbf{SymbolLKG} & \textbf{79.4} & 88.2 & \textbf{68.5} & \textbf{84.1} \\
\bottomrule

\end{tabular}
\end{sc}
\end{small}
\end{center}
\caption{Retrieval Performance (Recall@2 and Recall@5) on 2WikiMultiHopQA and HotpotQA. Baselines are categorized into Single-step and Multi-step approaches. SymbolLKG leverages topology-aware traversal to optimize evidence retrieval.}
\end{table*}

\paragraph{Results} \hyperref[tab:retrieval_results]{Table 3} presents the retrieval results. SymbolLKG attains $79.4$ / $88.2$ on 2WikiMultiHopQA and $68.5$ / $84.1$ on HotpotQA. The Recall@2 lead is driven by hybrid anchoring: three independently thresholded vector queries (over Rule, Constraint and Entity indices) are unioned with a BM25 entity-name pass, so the second slot is reliably occupied either by an entity-string hit or by a rule whose raw\_text contains the bridge. The HotpotQA gain further reflects the Logical Hull: two-hop traversal along [:MENTIONS], [:APPLIES\_TO] and typed predicate edges surfaces distractor-buried supporting facts that share no surface tokens with the query. The single trail, R@5 on 2Wiki, is consistent with our single-pass anchor-and-prune strategy, which maximises top-rank precision but cannot re-introduce evidence rejected by the initial threshold, unlike IRCoT's iterative re-querying.

\subsection{Question Answering Performance}
Finally, we assess the end-to-end question answering capabilities on 2WikiMultiHopQA and HotpotQA. This evaluation determines whether the retrieved context effectively supports the generation of accurate answers.

\paragraph{Baselines} We benchmark against three primary RAG frameworks: \textbf{NativeRAG}: A standard retrieval-augmented generation approach using dense retrieval. \textbf{GraphRAG} \cite{edge2024local}: A typical graph-enhanced RAG framework with structured knowledge modeling. \textbf{HippoRAG 2} \cite{gutierrez2025rag}: A state-of-the-art graph-based RAG method inspired by the hippocampal indexing theory.
\begin{table}[htbp]
    \centering
    \begin{tabular}{l|cc|cc|cc|cc}
    \toprule
    \multirow{2}{*}{\textbf{Method}} & \multicolumn{2}{c|}{\textbf{2Wiki}} & \multicolumn{2}{c|}{\textbf{HotpotQA}} & \multicolumn{2}{c|}{\textbf{Musique}} & \multicolumn{2}{c}{\textbf{Average}}\\
     & \textbf{EM} & \textbf{F1} & \textbf{EM} & \textbf{F1} & \textbf{EM} & \textbf{F1} & \textbf{EM} & \textbf{F1}\\
    \midrule
    NativeRAG   & 33.4 & 43.3 & 43.4 & 57.7 & 15.5 & 26.4 & 30.8 & 42.5\\
    GraphRAG    & 51.4 & 58.6 & 55.2 & 68.6 & 27.3 & 38.5 & 44.6 & 55.2 \\
    IRCoT + HippoRAG    & 47.7 & 62.7 & 45.7 & 59.2 & 21.9 & 33.3 & 38.4 & 51.7 \\
    HippoRAG 2 & 65.0 & 71.0 & 62.7 & 75.5 & 37.2 & 48.6 & 55.0 & 65.0 \\
    \midrule
    \textbf{SymbolLKG} & \textbf{70.2} & \textbf{74.9} & \textbf{73.8} & \textbf{81.5} & 
    \textbf{48.6} &
    \textbf{59.4} &
    \textbf{64.2} &  
    \textbf{71.9}   
      \\
    \bottomrule
    \end{tabular}
    \vspace{5pt}
    \caption{End-to-end QA Performance (EM and F1). We compare our SymbolLKG against baselines on 2WikiMultiHopQA, HotpotQA and Musique.}
    \label{tab:qa_results}
\end{table}

\paragraph{Metrics} We utilize \textbf{Exact Match (EM)} and \textbf{F1 Score}. EM measures whether the predicted answer $\hat{a}$ strictly matches the ground truth $a^*$ after normalization. F1 Score evaluates the token-level overlap. Let $T(\cdot)$ denote the bag of tokens function; precision ($P$) and recall ($R$) are computed as:
\begin{equation}
P = \frac{|T(\hat{a}) \cap T(a^*)|}{|T(\hat{a})|}, \quad R = \frac{|T(\hat{a}) \cap T(a^*)|}{|T(a^*)|}
\end{equation}
The F1 score is the harmonic mean: 
\begin{equation}
    \text{F1} = 2 \cdot (P \cdot R) / (P + R)
\end{equation}

\paragraph{Results} \hyperref[tab:qa_results]{Table 4} summarizes EM and F1 on the three multi-hop QA datasets. SymbolLKG attains $70.2$ / $74.9$ on 2Wiki, $73.8$ / $81.5$ on HotpotQA and $48.6$ / $59.4$ on MuSiQue, leading every prior baseline on every cell. The largest gain on MuSiQue reflects two complementary mechanisms: hash-based entity canonicalization merges paraphrased mentions of the same bridge entity (e.g., ``The Beatles'' and ``the band Beatles'') into a single graph node, so a bridge cannot be silently fragmented across single-hop sub-facts; meanwhile, the Logical Hull expansion admits rules and constraints two hops away from a question anchor even when they share no surface tokens with the query. The $+6.0$ F1 on HotpotQA reflects the same mechanism on noise rather than on composition: the topologically bounded subgraph isolates anchor-connected facts among 8--10 distractor paragraphs and is appended alongside the raw context, acting as a high-precision prior over which titles the answer LLM should attend to. The narrower gap on 2Wiki indicates that on cleaner Wikipedia text, dense retrieval baselines already cover most evidence, and the LKG's marginal contribution is mainly in entity disambiguation across same-surname or same-topic candidates.

\subsection{Router Performance}
\paragraph{Setup} To isolate the contribution of the Logic Router, we sample 100 instances from each of the five logical benchmarks ($N=500$ in total) and shuffle them into a single mixed evaluation stream. Each instance is processed end-to-end (LKG construction $\rightarrow$ hybrid retrieval $\rightarrow$ router decision) using Llama-3.3-70B-Instruct as the routing LLM. The ground-truth solver per dataset follows the design intent of the three engines: Z3 for AR-LSAT and LogicalDeduction (CSP / ordering / arithmetic); Prover9 for FOLIO (FOL with true/false/unknown) and ProofWriter, and Pyke for ProntoQA (Horn-clause forward chaining).

\paragraph{Results} \hyperref[tab:qa_results]{Table 5} reports the
per-dataset routing accuracy. SymbolLKG's router reaches an
overall accuracy of $86.0\%$, more than $2.5\times$ the random
baseline. Routing is essentially perfect on the two
Z3-targeted datasets ($100.0\%$ on both AR-LSAT and
LogicalDeduction), because the typed constraint subclasses
appear directly in the retrieved subgraph and trigger the
router's first decision rule. Routing is also strong on the two
Prover9-targeted datasets ($85.0\%$ on FOLIO, $92.0\%$ on
ProofWriter), where the three-valued cue ``true, false, or
uncertain/unknown'' in the question text reliably steers the
router to Prover9. The remaining error mass concentrates on
ProntoQA ($53.0\%$), where the dominant confusion is
Pyke-vs-Prover9: ProntoQA's ``true or false'' phrasing is
closed-world but lexically close enough to a Prover9 trigger to
occasionally mislead the router. This residual confusion is
largely benign in practice, however, because both engines
support Horn-clause forward chaining---an instance routed from
Pyke to Prover9 in this regime is still solved correctly more
often than not.

\begin{table}[htbp]
\centering
\small
\caption{Router classification accuracy (\%) on a mixed
evaluation set of 500 instances (100 per dataset). The
ground-truth solver per dataset is determined by the design
intent of each symbolic engine.}
\vspace{5pt}
\label{tab:router}
\begin{tabular}{lccc}
\toprule
Dataset & Ground-Truth Solver & Random & SymbolLKG \\
\midrule
AR-LSAT          & Z3      & 33.3 & \textbf{100.0} \\
LogicalDeduction & Z3      & 33.3 & \textbf{100.0} \\
FOLIO            & PROVER9 & 33.3 & \textbf{85.0}  \\
ProofWriter      & PROVER9 & 33.3 & \textbf{92.0}  \\
ProntoQA         & PYKE    & 33.3 & \textbf{53.0}  \\
\midrule
Average          & --      & 33.3 & \textbf{86.0}  \\
\bottomrule
\end{tabular}
\end{table}

\subsection{Computational Cost Analysis}

On $N=100$ samples per benchmark with Llama-3.3-70B-Instruct, the full SymbolLKG pipeline averages $122.2$ s per logical reasoning problem and $166$--$407$ s per multi-hop QA query, dominated by LLM API calls ($>\!99\%$ of wall-clock time). For deployments with fixed corpora the LKG is built once offline and reused across queries, reducing inference-time latency to retrieval ($<\!0.1$ s) plus a single answer-generation call ($\sim$$10$ s). Full per-stage breakdowns, distributions, and theoretical complexity are provided in Appendix~\ref{app:complexity}.

\section{Limitations \& Conclusion}

In this paper, we presented SymbolLKG, a neuro-symbolic framework that bridges the gap between the semantic flexibility of LLMs and the rigorous precision of symbolic solvers. By treating logical rules as topological nodes within a Logical Knowledge Graph and employing an adaptive router to dispatch tasks to the most appropriate engine (Z3, Prover9, or Pyke), our system significantly outperforms standard prompting and retrieval baselines on complex reasoning benchmarks. The core advantage of SymbolLKG lies in its \textbf{verifiability}: unlike the opaque generation process of Chain-of-Thought, our framework provides deterministic proof traces, ensuring that the reasoning process is both transparent and logically sound.

However, several limitations warrant further investigation. First is the \textbf{translation bottleneck}. Our framework operates
on the premise that the LLM can accurately parse natural language into formal graph representations. While the self-refining loop mitigates syntax errors, semantic misinterpretations during the extraction phase can still propagate to the solver. Essentially, the symbolic engine guarantees validity (the structure of the argument) but not soundness (the truth of the premises) if the initial grounding is flawed. Second, the \textbf{handling of ambiguity} remains a challenge. Formal logic requires precise definitions, whereas natural language is inherently ambiguous. When the input text contains vague quantifiers or metaphorical expressions, the rigid schema of the LKG may struggle to capture the full nuance, potentially leading to oversimplification. Finally, the introduction of graph construction and external solvers incurs additional \textbf{computational overhead}. While more accurate, our pipeline is computationally heavier than direct single-pass generation. Future work will focus on optimizing the extraction latency and exploring end-to-end differentiable neuro-symbolic alignment to reduce the dependency on intermediate discrete translation.

\newpage
\bibliographystyle{unsrt}
\bibliography{paper}

\newpage
\appendix

\section{Prompt Templates}
Appendix A provides the specific prompt templates utilized within the SymbolLKG framework to facilitate reproducibility and illustrate the detailed interaction mechanisms between the system and the Large Language Model.

\subsection{Prompt for LKG Construction}
\begin{figure}[ht!]
    \centering
    \includegraphics[width=\textwidth]{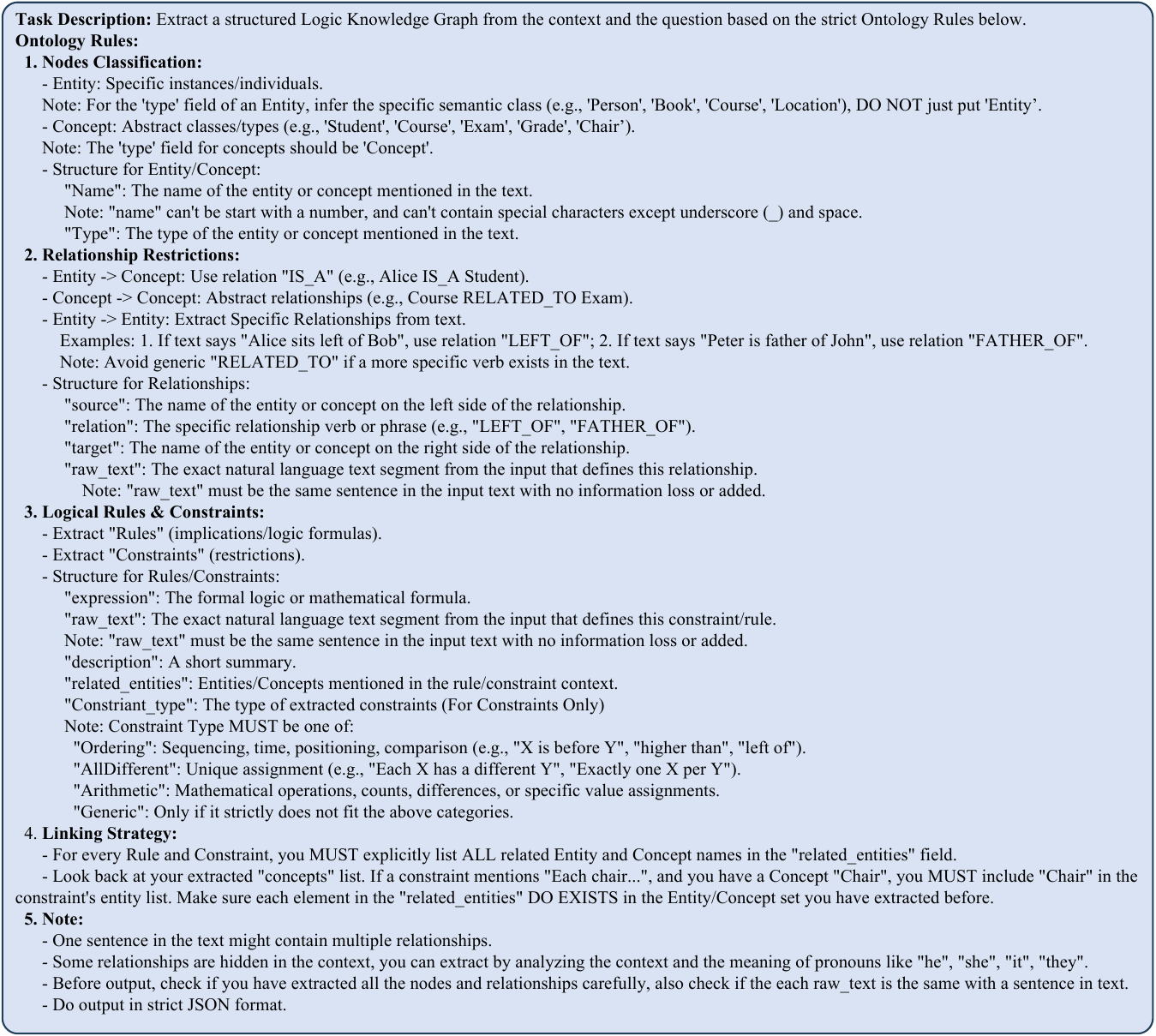}
    \caption{Prompt template for Logical Knowledge Graph Construction. This prompt instructs the LLM to extract structured logic from text based on a dynamic ontology. It defines specific schemas for Entities, Concepts, and logical relationships (Rules/Constraints), enforcing a strict JSON output format to ensure graph integrity.}
    \label{fig:LKG_Construction}
\end{figure}
\newpage

\subsection{Prompt for Solvers}
\begin{figure}[ht!]
    \centering
    \includegraphics[width=\textwidth]{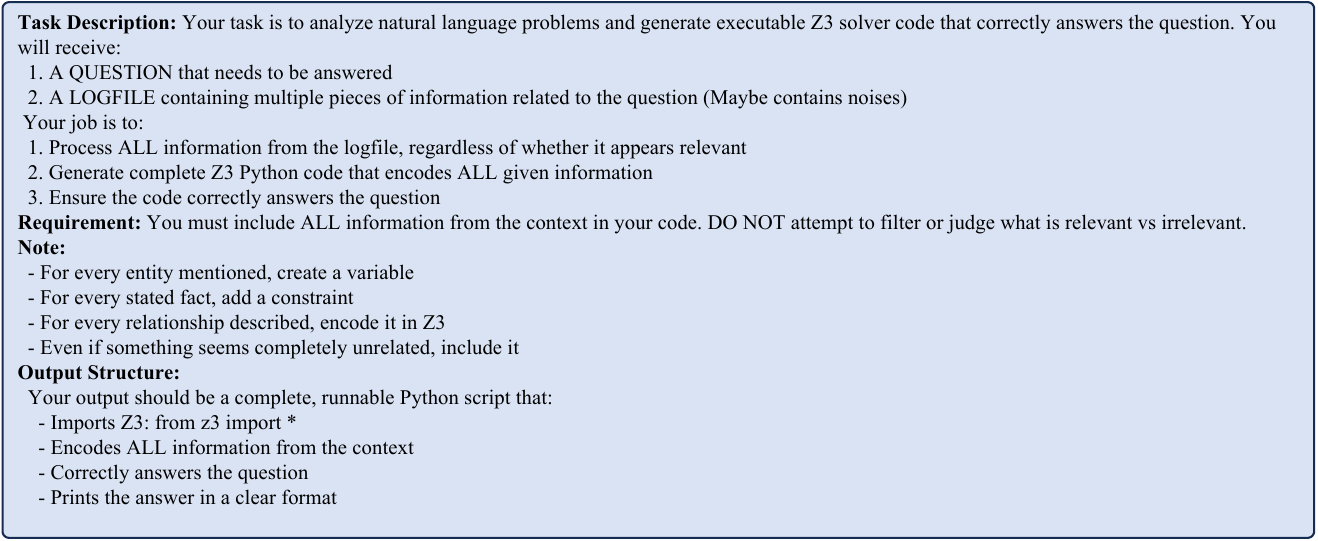}
\end{figure}

\begin{figure}[ht!]
    \centering
    \includegraphics[width=\textwidth]{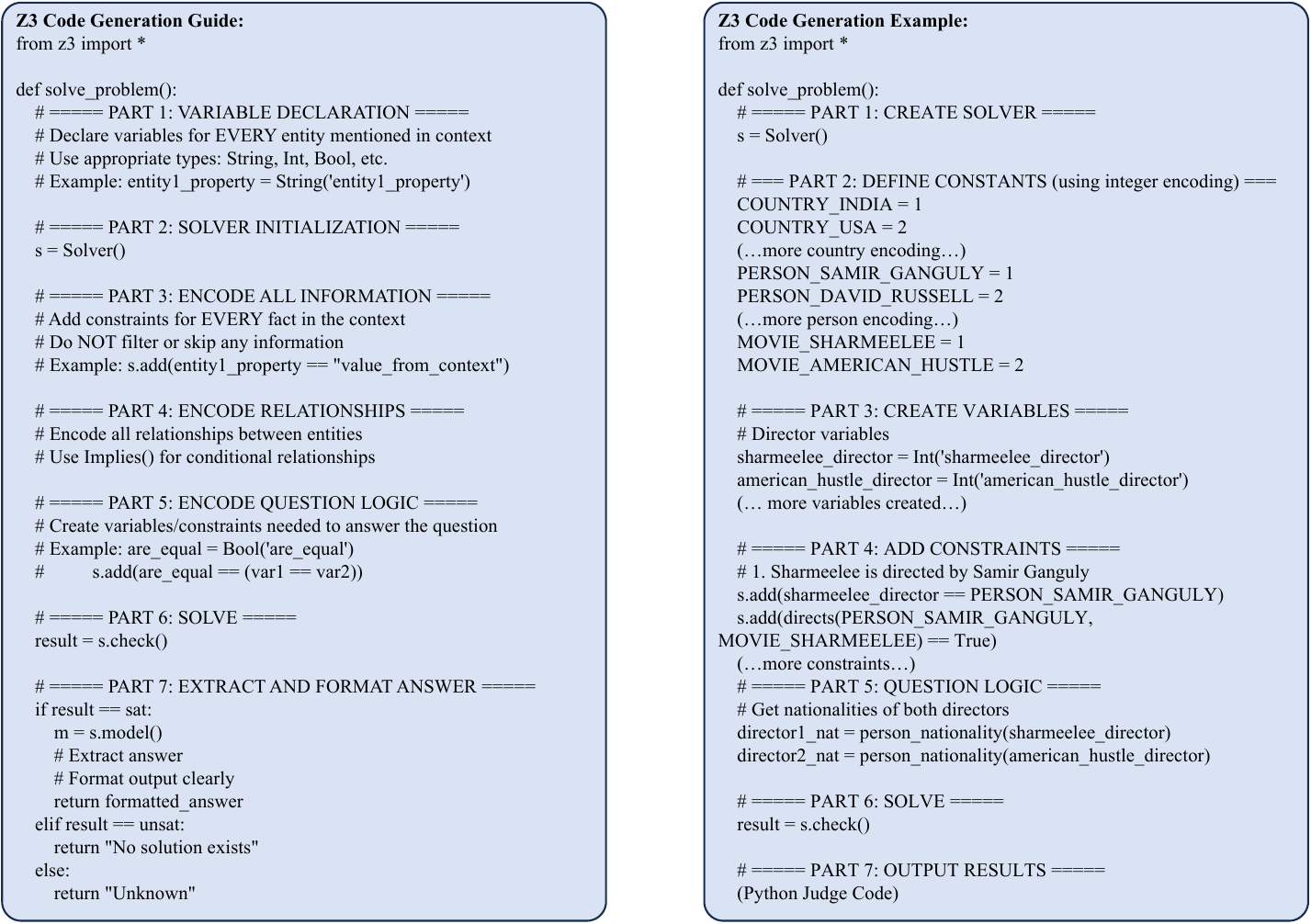}
    \caption{Prompt template for the Symbolic Solver (Z3) Code Generation. This prompt guides the model to translate the retrieved logical subgraph into an executable Python script utilizing the Z3 SMT solver. It enforces a structured coding process: declaring variables for entities, encoding facts as constraints, and formulating the specific query logic to verify the answer.}
    \label{fig:Solver}
\end{figure}

\newpage
\subsection{Prompt for Pruner}

\begin{figure}[ht!]
    \centering
    \includegraphics[width=\textwidth]{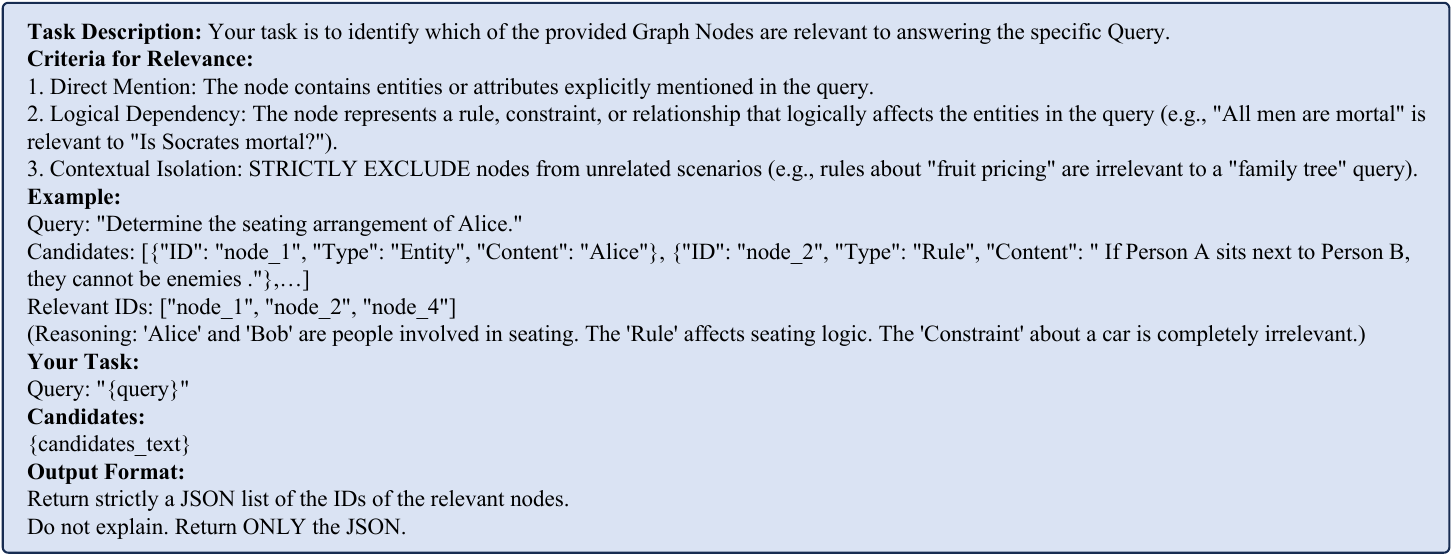}
    \caption{Prompt template for the Logic Pruning Module. This prompt instructs the LLM to filter the expanded candidate set. }
    \label{fig:pruner}
\end{figure}

\subsection{Prompt for Router}
\begin{figure}[ht!]
    \centering
    \includegraphics[width=\textwidth]{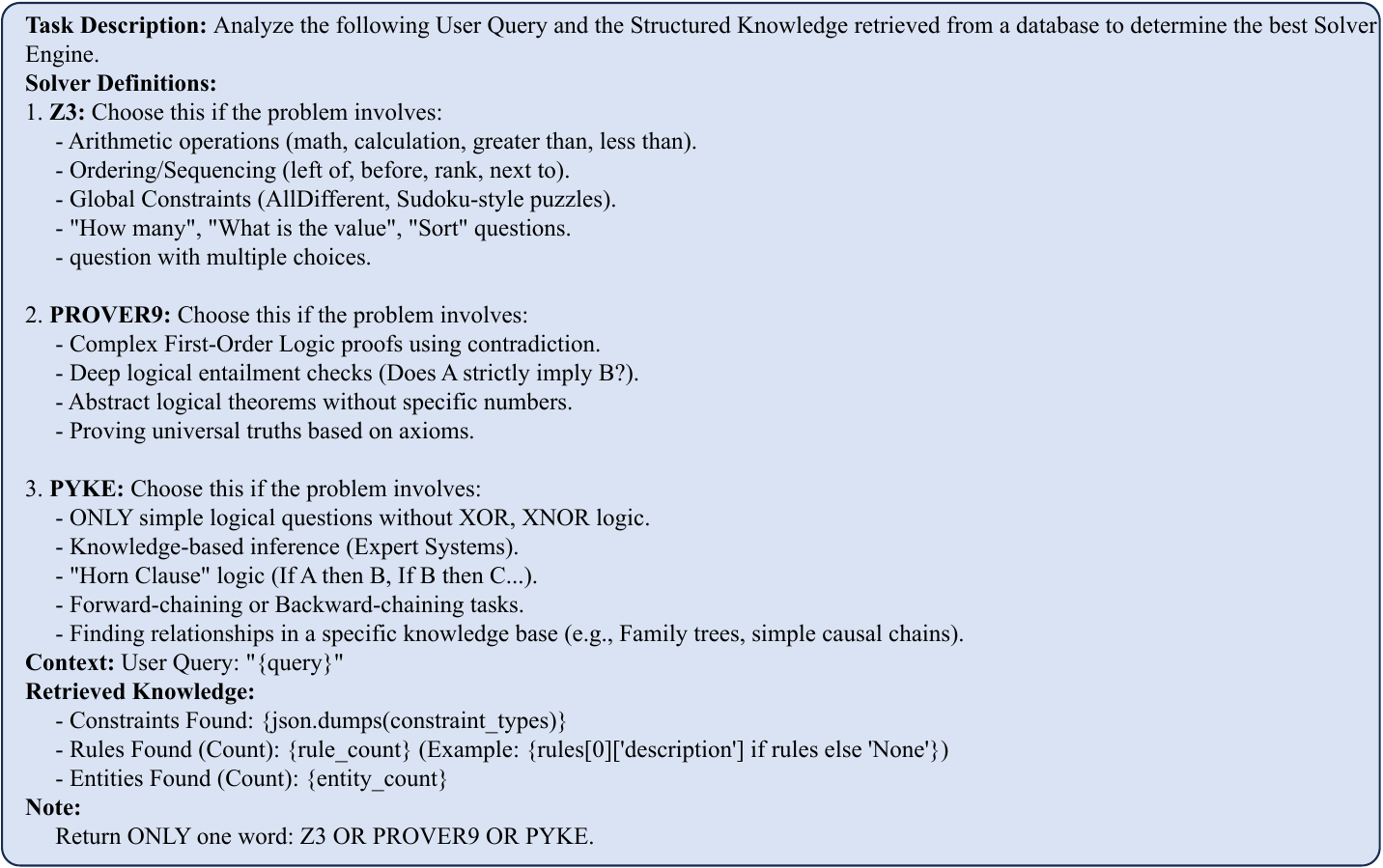}
    \caption{Prompt template for the Adaptive Logic Router. This prompt guides the model to classify the reasoning task by evaluating the retrieved constraints and rules. It explicitly defines the decision boundaries for solver selection: Z3 is chosen for arithmetic and constraint satisfaction problems (CSPs), Prover9 for first-order logic theorem proving, and Pyke for knowledge-based relational inference.}
    \label{fig:router}
\end{figure}
\newpage
\section{Case Study}
Appendix B presents a comprehensive case study derived from the AR-LSAT benchmark to demonstrate the end-to-end workflow of the SymbolLKG framework. It demonstrates the complete lifecycle of a logical reasoning task, detailing the intermediate outputs at each stage: from the initial extraction of the Logical Knowledge Graph (LKG) components (Entities, Concepts, Rules, and Constraints) to the dynamic inference process involving hybrid retrieval, logic routing, and the final generation of executable Z3 symbolic code.
\begin{figure}[ht!]
    \centering
    \includegraphics[width=\textwidth]{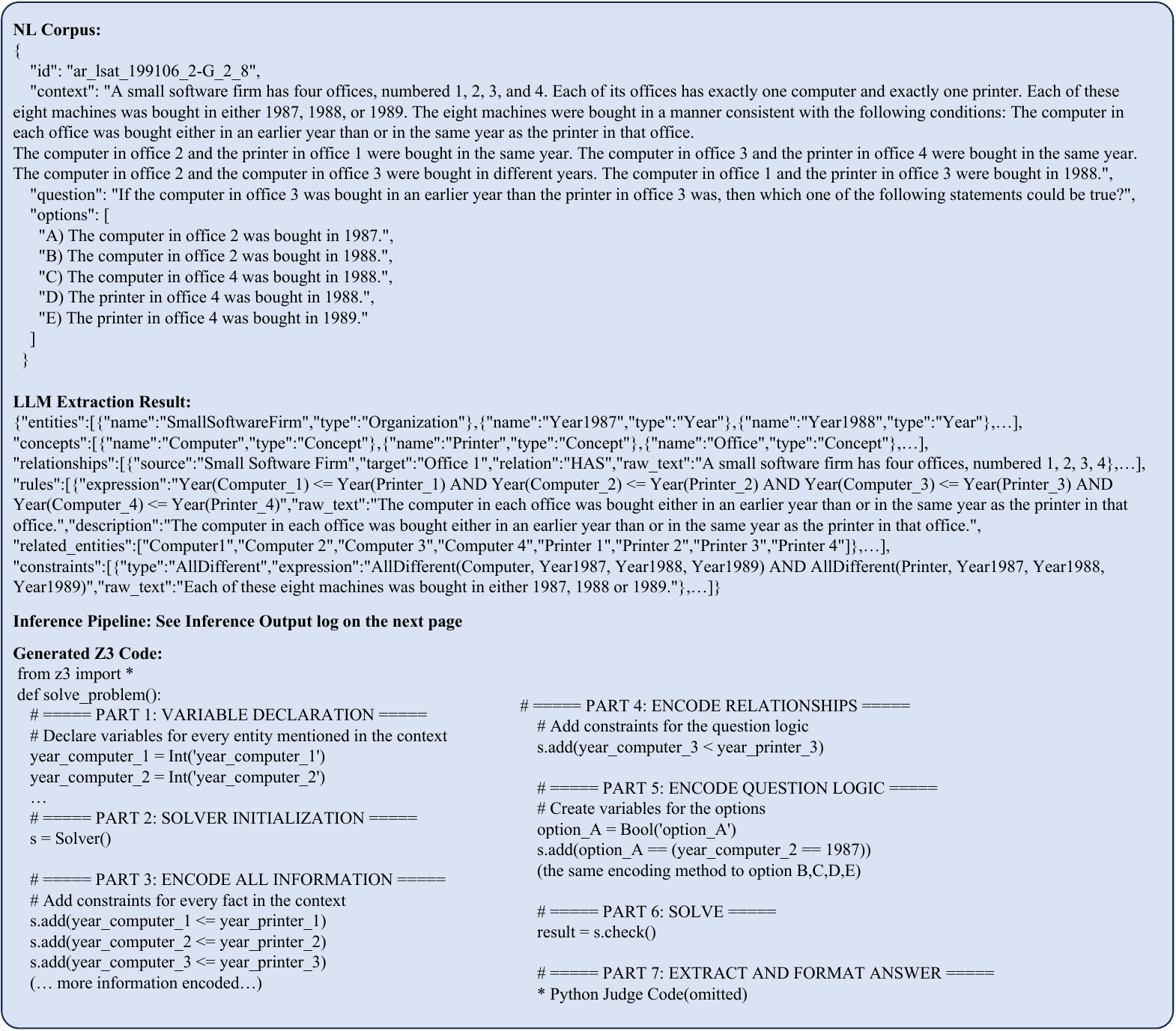}
    \caption{Case Study: Logical Knowledge Graph Extraction Result. The figure displays the structured JSON output extracted from an AR-LSAT logical reasoning problem. It demonstrates how the system identifies entities (e.g., Office, Computer), maps concepts, and formalizes complex constraints (e.g., ``AllDifferent'', ``Ordering'') and logical rules from the raw text context.}
    \label{fig:casestudy1}
\end{figure}
\clearpage
\begin{figure}[htbp!]
    \includegraphics[width=\textwidth]{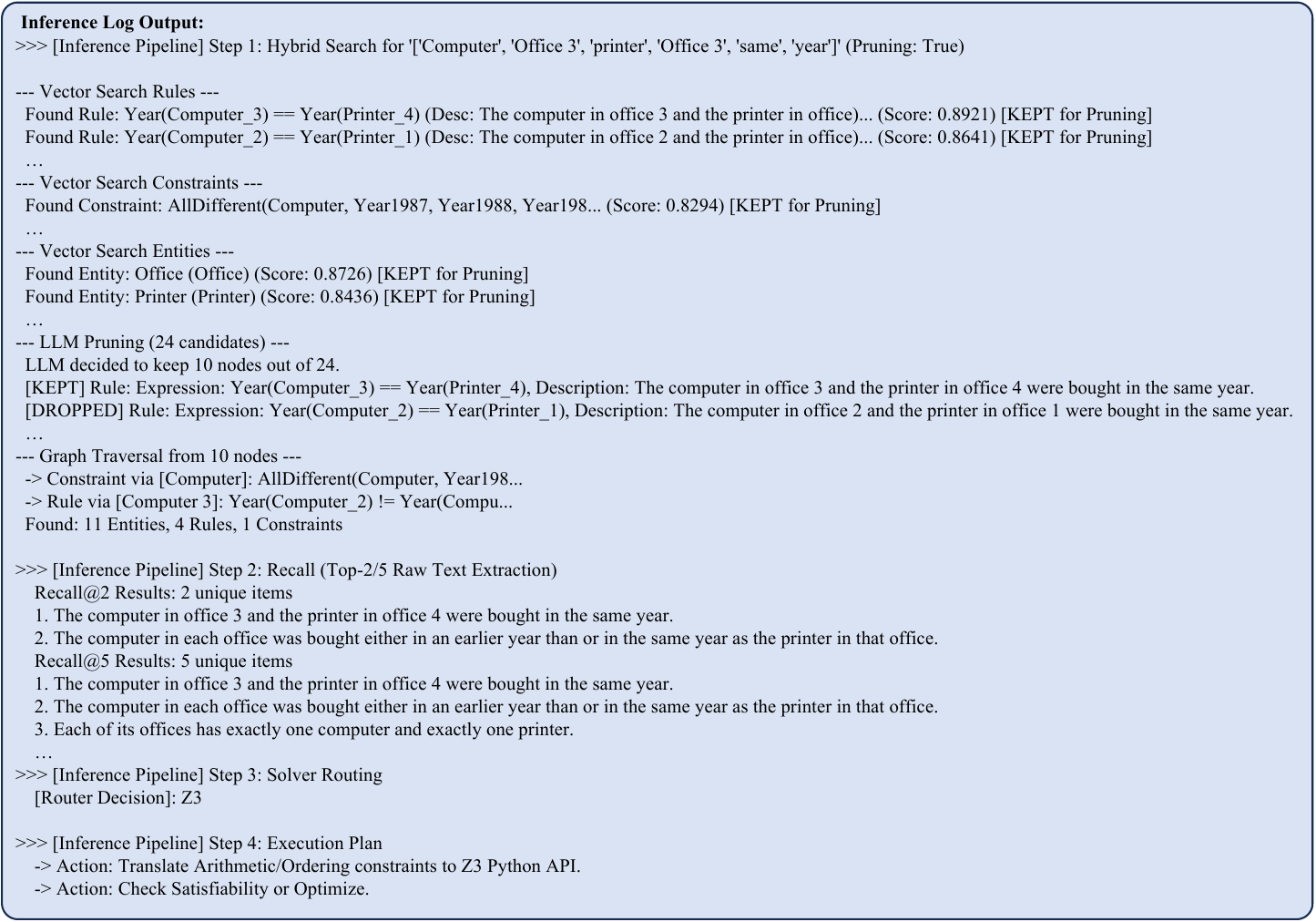}
    \caption{Case Study: Inference Pipeline Log. This log trace illustrates the step-by-step execution flow: (1) Hybrid Search retrieves relevant rules regarding ``Year'' and ``Office''; (2) Pruning filters candidates down to the essential logic; (3) The Logic Router correctly identifies the task as an arithmetic/ordering problem and selects the Z3 Solver; (4) The system generates and executes valid Z3 code to derive the final answer.}
    \label{fig:casestudy2}
\end{figure}

\section{Time Complexity Analysis}
\label{app:complexity}

\subsection{Empirical Latency}
We profile per-sample wall-clock time on $N=100$ instances per
benchmark.
\paragraph{Multi-hop QA}
Table~\ref{tab:time_qa} reports the per-sample latency of the
full SymbolLKG pipeline on the three multi-hop QA datasets.
End-to-end latency ranges from $166.0$ s on 2WikiMultiHopQA to
$407.1$ s on MuSiQue, with the bulk of the cost
($159.2$--$395.8$ s) spent on per-question LKG construction.
The LKG itself triggers entity, rule, and constraint extraction
over the question's $8$--$10$ paragraph context; the
answer-generation call afterwards takes only $6.0$--$11.4$ s.
\begin{table}[htbp]
\centering
\small
\caption{Per-sample wall-clock latency (seconds) on the
multi-hop QA pipeline, averaged over $N=100$ instances per
benchmark.}
\label{tab:time_qa}
\begin{tabular}{lrrr}
\toprule
Dataset    & LKG    & LLM   & Total  \\
\midrule
2Wiki      & 159.2  & 6.79  & 165.96 \\
HotpotQA   & 208.5  & 6.00  & 214.57 \\
MuSiQue    & 395.8  & 11.35 & 407.14 \\
\bottomrule
\end{tabular}
\end{table}
\paragraph{Logical Reasoning}
Table~\ref{tab:time_logic} reports the per-sample latency of
the logical reasoning pipeline (single-path: LKG construction
plus the routed-solver path including self-refining retries).
End-to-end latency averages $122.2$ s per problem (LKG
construction $29.0$ s, routed solver $93.2$ s). ProofWriter
and AR-LSAT exhibit the longest mean times ($166.5$ s and
$135.6$ s respectively), reflecting their longer
Prover9 / Z3 programs and higher retry rates; ProntoQA is
shortest ($71.4$ s) thanks to the lightweight Pyke
forward-chaining path.
\begin{table}[htbp]
\centering
\small
\caption{Per-sample wall-clock latency (seconds) on the
logical reasoning pipeline, averaged over 20 instances per
dataset ($N=100$ in total). ``Total'' reflects single-path
production deployment (LKG construction plus the routed-solver
path with up to three self-refining retries).}
\label{tab:time_logic}
\begin{tabular}{lrrr}
\toprule
Dataset          & LKG  & Solver & Total \\
\midrule
AR-LSAT          & 33.1 & 102.5  & 135.6 \\
LogicalDeduction & 18.9 & 103.3  & 122.2 \\
FOLIO            & 23.4 &  91.8  & 115.2 \\
ProntoQA         & 34.3 &  37.1  &  71.4 \\
ProofWriter      & 35.1 & 131.4  & 166.5 \\
\midrule
Average          & 29.0 &  93.2  & 122.2 \\
\bottomrule
\end{tabular}
\end{table}
\subsection{Discussion}
The dominant cost is LLM API latency rather than graph
operations: across both pipelines, the two recurring LLM calls
(LKG extraction plus solver code generation or answer
generation) jointly account for over $99\%$ of wall-clock
time, while retrieval and BFS combined remain below $0.1$ s
per query. Three properties of the framework limit the
practical impact of this overhead. (i)~For fixed corpora
served at deployment time, the LKG can be built once offline
and reused across all queries, eliminating the
$155$--$494$ s per-question construction cost that dominates
the multi-hop QA numbers; under this amortisation, the
inference-time cost reduces to retrieval ($<\!0.1$ s) plus a
single answer-generation call ($\sim$$10$ s). (ii)~The
self-refining loop is bounded at three retries, so worst-case
solver latency is at most $3\times$ the single-attempt cost.
(iii)~The routing and pruning calls operate on bounded-length
prompts and can be served by a smaller, cheaper backbone
without materially affecting accuracy. The remaining overhead
reflects the cost of trading probabilistic single-pass
inference for verifiable symbolic reasoning.


\end{document}